\documentclass[11pt,a4paper]{article}
\usepackage{etoolbox}
\usepackage[utf8]{inputenc}
\makeatletter
\patchcmd\@combinedblfloats{\box\@outputbox}{\unvbox\@outputbox}{}{\errmessage{\noexpand patch failed}}
\makeatother

\usepackage[hyperref]{naaclhlt2019}
\usepackage{times}
\usepackage{latexsym}
\usepackage{amsmath}
\usepackage{amssymb}
\usepackage{amsthm}
\usepackage{statistics-macros}
\usepackage{algorithm}
\usepackage{multirow}
\usepackage{booktabs}
\usepackage{tabularx}
\usepackage{algorithmic}
\usepackage{tablefootnote}

\newtheorem{prop}{Proposition}
\newtheorem{thm}{Theorem}

\usepackage{todonotes}

\usepackage{url}

\aclfinalcopy %

\newcommand{\tvhp}{\text{HUSE}}
\newcommand{\tvh}{\text{HUSE-Q}}
\newcommand{\tvdiv}{\text{HUSE-D}}

\newcommand{\phum}{p_{\text{ref}}}

\newcommand{\pmodel}{p_{\text{model}}}
\newcommand{\hj}{\text{HJ}}
\newcommand{\len}{\text{len}}

\newcommand{\optclass}{L^*}

\newcommand{\fhum}{f_\text{hum}}
\newcommand{\phiopt}{\phi_\text{opt}}
\newcommand{\phihuse}{\phi_\text{huse}}
\newcommand{\phihj}{\phi_\text{hj}}

\newcommand\phibrack[1]{\left[#1\right]}

\title{Unifying Human and Statistical Evaluation for \\
  Natural Language Generation }

 \author{Tatsunori B. Hashimoto*$^{1,2}$ ~~ Hugh Zhang*$^1$ ~~ Percy Liang$^{1,2}$ \\
   \textbf{(* equal contribution)} \\
   $^1$Department of Computer Science ~~ $^2$Department of Statistics \\
   Stanford University \\
   {\small \tt \{thashim,hughz\}@stanford.edu ~~ pliang@cs.stanford.edu} \\
 }

\date{}

\begin{document}
\maketitle
\begin{abstract}
How can we measure whether a natural language generation system produces both high quality and diverse outputs?
Human evaluation captures quality but not diversity, as it does not catch models that simply plagiarize from the training set.
On the other hand, statistical evaluation (i.e., perplexity) captures diversity but not quality,
as models that occasionally emit low quality samples would be insufficiently penalized.
In this paper, we propose a unified framework
which evaluates both diversity and quality, based on the optimal error rate of predicting whether a sentence is human- or machine-generated.
We demonstrate that this error rate can be efficiently estimated by combining human and statistical evaluation, using an evaluation metric which we call HUSE. 
On summarization and chit-chat dialogue,
we show that (i) HUSE detects diversity defects which fool pure human evaluation
and that (ii) techniques such as annealing for improving quality actually decrease HUSE due to decreased diversity.

\end{abstract}

\section{Introduction}

Generating text is a core part of many NLP tasks such as
image captioning \citep{lin2014microsoft},
open-domain dialogue \cite{sordoni2015neural},
story generation \cite{roemmele2016writing},
and summarization \cite{nallapati2016abstractive}.
However, proper evaluation of natural language generation has proven difficult \citep{liu2016evaluate,novikova2017why,chaganty2018evaluation}.
A good evaluation metric should not only capture the \emph{quality} of generation,
but also the \emph{diversity} of generation,
which is especially crucial for creative, open-ended tasks like dialogue or story generation.

\emph{Human evaluation}, which is often viewed as the gold standard evaluation,
 captures quality but fails to capture diversity.
As an example, for language modeling, a model that directly plagiarizes sentences
from the training set would pass the human quality bar
but would have zero generalization ability and thus have inadequate diversity.
On the other hand, \emph{statistical evaluation}---i.e., perplexity on a reference test set---captures diversity,
as it ensures a model must assign reasonable probability to novel sentences,
but perplexity provides an inadequate measure of quality \citep{theis2015note}. For example, modifying a perfect model by removing its ability to generate even a single test sentence results in \emph{infinite} perplexity even though the model is still near-perfect.
Automatic metrics such as BLEU \citep{papineni02bleu} and ROUGE \citep{lin2004rouge}
capture quality better than perplexity but still correlate poorly with human evaluation and fail to capture diversity
\cite{novikova2017why,chaganty2018evaluation}.

\begin{figure}
  \includegraphics[scale=0.54]{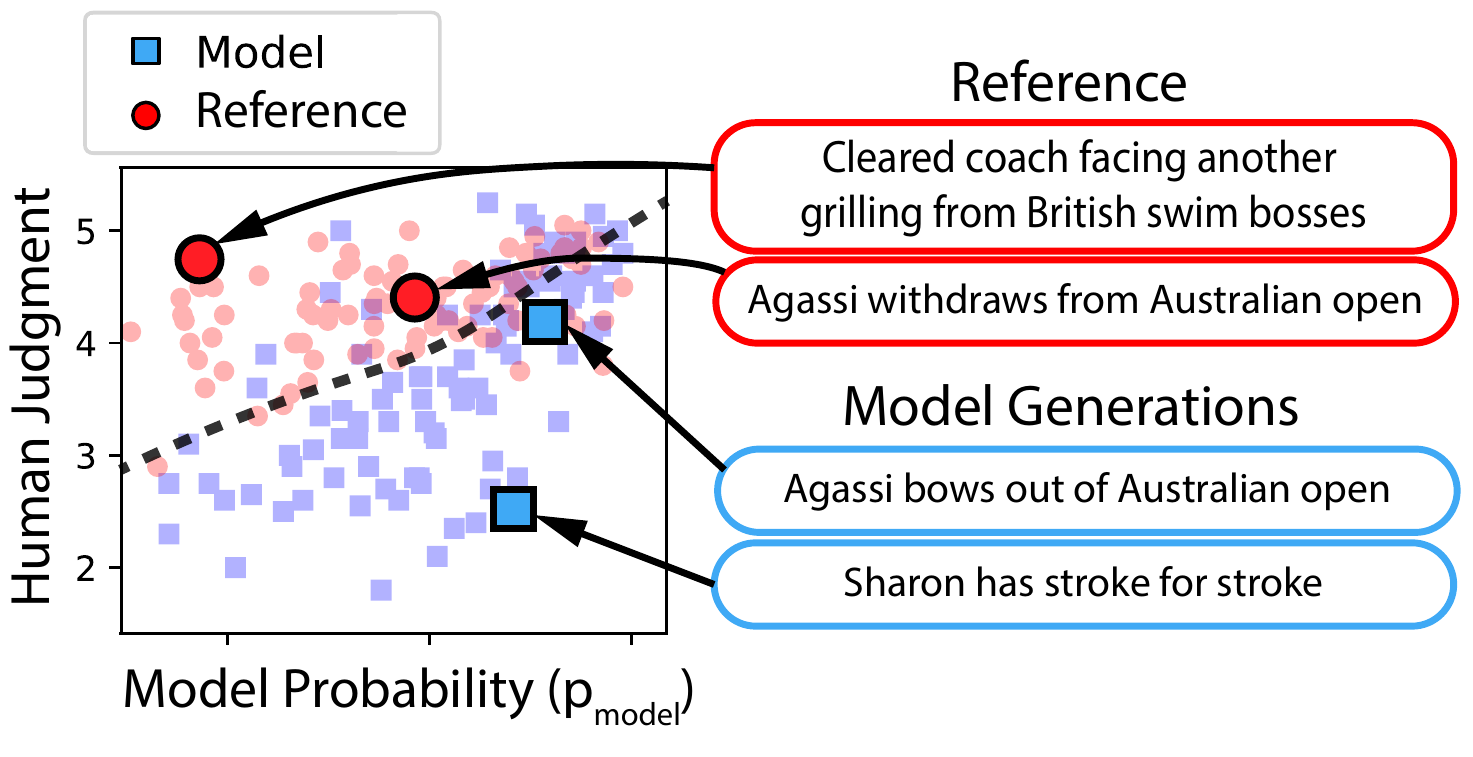}
  \caption{HUSE is twice the classification error of distinguishing reference and generated text based on human judgment scores and model probabilities.
  HUSE identifies samples with defects in quality (Sharon has stroke $\hdots$) and diversity (Cleared coach facing $\hdots$).
  }
  \label{fig:figone}
  \end{figure}

Existing approaches to combining statistical and human evaluation have been ad-hoc, leading to misleading performance measures. A common approach is to measure diversity through the perplexity of a probabilistic model and quality through human evaluation on beam-searched outputs. This gives the illusion that a single model is high-quality and diverse, while the reality is that it shows we can have either a diverse model (when sampling from the distribution used to compute perplexity) \emph{or} a high-quality model (when beam-searching).

In this paper, we define the idealized evaluation metric as
twice the error of the \emph{optimal discriminator}
for classifying sentences as coming from the reference distribution or the model (Section~\ref{sec:optimal}).
If a model generates gibberish (low quality), the optimal discriminator can classify these accurately as coming from the model.
If the reference distribution contains sentences the model cannot generate (low diversity),
the optimal discriminator can classify these accurately as coming from the reference.

Unfortunately, the optimal discriminator is unavailable.
Human discriminators cannot capture diversity effectively,
and learned discriminators---e.g., from a Generative Adversarial Network \citep{goodfellow2014gan}
or one trained on human judgments \citep{lowe2017towards}---are too unreliable to use for rigorous evaluation.

Our key result (Section~\ref{sec:huse}) is based on the observation that the optimal classifier depends only on two numbers:
the probability of a sentence under the model
and the probability under the reference distribution.
The former can be computed directly from the model, and we show that the latter can be well-approximated by human judgment scores.
The resulting two-dimensional space is illustrated in Figure \ref{fig:figone}.
We apply a simple $k$-nearest neighbor classifier in this space and
define Human Unified with Statistical Evaluation (HUSE) as twice the leave-one-out error of this classifier.

We apply $\tvhp$ to four natural language generation tasks (Section~\ref{sec:experiments}):
language modeling, chitchat dialogue, story generation, and summarization.
First, we show that human evaluation alone is insufficient to discriminate model generations from the references, leading to inflated estimates of model performance.
In contrast, \tvhp{} is able to reveal deficiencies of current models.
We also show that common techniques for improving sample quality such as annealing
actually increase distinguishability between the model and reference due to losses in diversity.

\section{Optimal Discriminator}
\label{sec:optimal}

Consider a natural language generation task where the model is given a context $x$ (e.g., a dialogue history) drawn from some prior $p(x)$
and must output a distribution over possible sentences $\pmodel(y \mid x)$.
We define an idealized evaluation metric based on whether $\pmodel$ is close to a \emph{reference distribution} $\phum$,
which is generally human-generated.\footnote{
While some tasks only care about quality and thus only require $\pmodel$ to place mass on \emph{some} high quality $y$,
we demand that $\pmodel$ places mass on \emph{all} high quality $y$ as given by $\phum$.
This diversity is important for open-ended tasks such as dialogue or story generation.
Also note that $\phum$ need not be the human distribution, or match the training distribution. It can be defined as the distribution given by experts.
}
Specifically, consider a random variable $y$ drawn from
either the reference or the model based on an indicator $z \sim \text{Bernoulli}\left(\frac{1}{2}\right)$:
\begin{align}
y \mid x, z \sim
\begin{cases}
  \phum(y \mid x)   & \text{ if } z=1 \\
  \pmodel(y \mid x) & \text{ if } z=0.
\end{cases}
\end{align}
Define $\optclass$ to be twice the lowest possible error over any discriminator $f$
that attempts to determine $z$ based on $x$ and $y$:
\begin{align}
\label{eqn:optimal}
\optclass \defeq 2 \inf_f \P[f(x, y) \neq z].
\end{align}
$\optclass$ measures similarity between $\pmodel$ and $\phum$;
it is 0 if $\pmodel$ and $\phum$ are disjoint and 1 if they are identical.\footnote{
Note that $\optclass$ is a linear function of the total variational divergence:
$\|\pmodel - \phum\|_\text{TV} \defeq \sum_{x,y} p(x) \left|\pmodel(y \mid x) - \phum(y \mid x)\right| = 1 - \optclass$. See Appendix \ref{sec:tverr} for details.}

\paragraph{Obstacles.}

Unfortunately, $\optclass$ is unattainable because it requires
computing the optimal discriminator.
In the spirit of the Turing Test, we could consider using the error rate of a human discriminator $\fhum$ instead,
often considered the gold standard for evaluation.
However, while humans might have knowledge of $\phum$, they do not have full knowledge of $\pmodel$ and thus would have difficulties determining which sentences a model \emph{cannot} generate.

As a concrete example, suppose $\phum$ placed a uniform distribution over some set $S$.
Without knowledge of $\pmodel$ the most sensible discriminator is to predict $z=1$ (reference)
when $y \in S$.
This discriminator achieves the same classification error of $0.5$ for both the perfect model $\pmodel = \phum$ and one which can only return a single $y \in S$. We could try to reveal $\pmodel$ to humans by showing multiple samples simultaneously, but this is expensive
and, as we will later see, unnecessary.

Another option is to learn $f$ over an expressive class of
functions such as neural networks on data sampled from $\pmodel$ and $\phum$.
This is analogous to learning the discriminator in a
Generative Adversarial Network (GAN) \citep{goodfellow2014gan} or
learning an evaluation metric from human judgments \citep{lowe2017towards}.
However, as $(x,y)$ are high-dimensional objects,
training a good classifier is extremely difficult (and perhaps not significantly easier
than solving the original generation problem).
Indeed, learned evaluation metrics do not generalize very well \citep{lowe2017towards,chaganty2018evaluation}.
Unlike these approaches which seek to replace human evaluation, our focus will
instead be on combining human and automatic statistical evaluation to estimate the \emph{optimal} classifier error.

\section{Human Unified with Statistical Evaluation (HUSE)}
\label{sec:huse}

Our key result is that the optimal discriminator depends on $(x,y)$ only
through a two-dimensional sufficient statistic (Section~\ref{sec:twoSuffice}),
motivating an approximation which we call \tvhp{} (Section~\ref{sec:humanPhi}).

For any feature map $\phi$ that maps $(x,y)$ to $\phi(x,y) \in \R^d$,
define the evaluation score $L(\phi)$ to be twice the error rate of the optimal discriminator that
depends on $(x,y)$ \emph{only} through $\phi$:
\begin{align}
  \label{eqn:Lphi}
  L(\phi) \defeq 2 \inf_f \P[f(\phi(x, y)) \neq z].
\end{align}

Note that the evaluation score $L(\phi)$ given by a feature map $\phi$
optimizes over all functions that depend on $\phi$ (\ref{eqn:Lphi}).
Thus, the more information $\phi$ contains, the lower $L(\phi)$ is.
This has two implications:
First, any feature map $\phi$ yields an (optimistic) \emph{upper bound} on $L^*$ (\ref{eqn:optimal}),
meaning that $L(\phi)$ might be able detect when a model is poor but cannot certify that it is good.
Second, adding features to $\phi$ can only improve this bound.

\subsection{Two features suffice}
\label{sec:twoSuffice}

Let us consider the following two-dimensional feature map:
\begin{align}
  \phiopt(x,y) \eqdef \phibrack{\phum(y \mid x), \pmodel(y\mid x)}.
\end{align}
From the arguments above, it is clear that $L(\phiopt) \ge \optclass$,
but perhaps more surprisingly, we actually have equality:
\begin{proposition}\label{prop:phiopt}
The two-dimensional feature map $\phiopt$ achieves the optimal discriminator score:
$L(\phiopt) = \optclass$.
\end{proposition}
\begin{proof}
We compute the true posterior over $z$ given $x,y$.
  Since $p(z = 1) = p(z = 0) = \frac12$,
  $p(y \mid x, z = 1) = \phum(y \mid x)$ and
  $p(y \mid x, z = 0) = \pmodel(y \mid x)$,
  by Bayes' rule:
\begin{align*}
  \label{eqn:bayes}
  p(z = 1 \mid x, y) = \frac{\phum(y \mid x)}{\phum(y \mid x) + \pmodel(y \mid x)}.
\end{align*}
  The optimal discriminator simply predicts $z = 1$ if $\phum(y \mid x) > \pmodel(y \mid x)$ and $z = 0$ otherwise.
  In other words, the decision boundary is given by $\phiopt(x,y)_1 > \phiopt(x,y)_2$.
\end{proof}
More generally, we can obtain this equality with a wider class of $\phi$. It will hold exactly for any invertible transformation of $\phiopt$ (Appendix Corollary \ref{cor:invertexact}), and approximately for any $\phi$ which has high mutual information with $\phiopt$ (Appendix Theorem \ref{thm:approxerr}). This means that we can substitute $\phum$ with noisy, possibly un-normalized estimates and still obtain accurate estimates of $\optclass$.

\subsection{HUSE features}
\label{sec:humanPhi}

\begin{figure}[h!]
\centering
\includegraphics[scale=0.35]{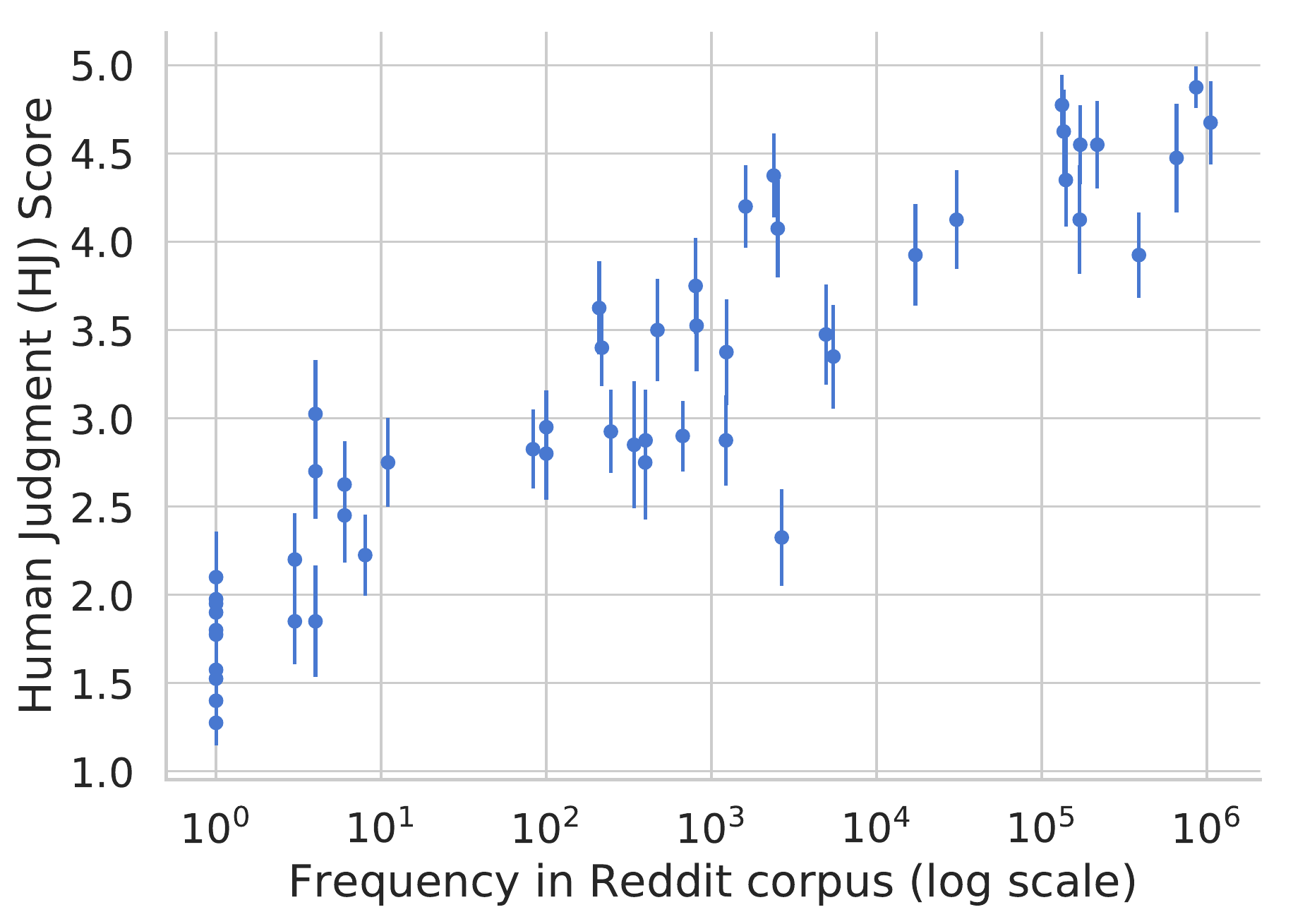}
\caption{
On the Reddit corpus,
  human judgment ($\hj$) of the ``typicality'' of a sentence $y$ correlates strongly ($r = 0.92$) with its frequency in the corpus,
  suggesting that $\hj$ is a good surrogate for $\log \phum$.
Error bars at the 90\% confidence interval.
}
\label{fig:redfreq}
\end{figure}

While we can directly compute $\pmodel(y \mid x)$ for many probabilistic models,
$\phum(y \mid x)$ is unattainable, so $L(\phiopt)$ is not computable.
However, the \emph{wisdom of the crowds} \cite{surowiecki2004wisdom, ungar2012judgement} suggests that pooling
together the judgments of many humans
can often produce surprisingly reliable estimates of real-world probabilities such as $\phum(y\mid x)$,
even if no individual human is particularly reliable.
With this motivation, we ask Amazon Mechanical Turk workers to rate a sentence from 1--5 based on how ``typical'' it is as a way to estimate $\phum(y\mid x)$.
(see Appendix~\ref{sec:amt} for more details). 
We define $\hj(x, y)$ to be the average response over 20 crowdworkers.
Figure~\ref{fig:redfreq} shows that for a language modeling task on the Reddit corpus,\footnote{We used the Reddit corpus due to crowdworker
familiarity, corpus size, and short average sentence length, which results in a wide range of sentence frequencies.}
$\hj(x,y)$ strongly correlates with the actual log-frequency of $y$ in the corpus.
The high correlation suggests that human judgments $\hj(x,y)$ are a good surrogate for $\log\phum$.

In addition, we found that rather than using the model probability $\pmodel(y \mid x)$ directly
as a feature, normalizing by sentence length $\len(y)$ yielded lower (tighter) scores.
We therefore define the HUSE features as follows:
\begin{align}
  \phihuse(x,y) \eqdef \phibrack{\frac{\log \pmodel(y \mid x)}{\len(y)}, \hj(x, y)},
\end{align}
and define the (population) \emph{HUSE score} as $L(\phihuse)$.

\subsection{Guarantees derived from HUSE}
\label{sec:properties}

We now show that the HUSE score satisfies two nice properties: (i) HUSE does at least as well as human evaluation and (ii) a low HUSE score is \emph{sufficient} to show that a model is far from the reference distribution.

To show (i), consider a feature map that only includes human evaluation: $\phihj(x,y) \eqdef [\hj(x,y)]$.
Because $\phihuse$ also incorporates human evaluation,
$L(\phihuse)$ is always tighter (lower) than the human discriminator error $L(\phihj)$:
\begin{prop}[Relationship between HUSE, human evaluation, and optimal scores]
\label{prop:rel}
\begin{align}
L(\phihj) \geq L(\phihuse) \geq \optclass.
\end{align}
\end{prop}
Furthermore, the main difference between $L(\phihuse)$ and $\optclass$ is that
the former uses $\hj(x,y)$ and the latter uses $\phum$.
But as we argued using Figure~\ref{fig:redfreq},
$\hj(x, y)$ is strongly correlated with $\phum$, and good approximations to $\phum$ provide approximation guarantees for $L(\phihuse)$ (Appendix Theorem \ref{thm:approxerr}).

\section{Evaluating models with HUSE}
\label{sec:estimation}

In this section, we show how we can estimate the error rate $L(\phi)$ from finite data (Section \ref{sec:estopt}).
We then show how the HUSE estimate $(\hat{L}(\phihuse))$ can be decomposed
into a score that measures quality (HUSE-Q)
and a score that measures diversity (HUSE-D),
which allows us to study quality-diversity tradeoffs (Section \ref{sec:moddef}).

\subsection{Learning a discriminator}
\label{sec:estopt}

For any feature map $\phi$, we show how to produce an estimate of $L(\phi)$.
Fix $n$ contexts $x_1, \dots, x_n$.
First, we draw $n$ examples $y_1, \dots, y_n$ from the reference distribution $\phum(y \mid x)$,
which are usually human-generated sentences from a test set.
We also draw $n$ examples $y_1', \dots, y_n'$ from the model $\pmodel(y \mid x)$ we wish to evaluate.
Next, for each of the $2n$ examples $(x, y)$, we compute the feature map $\phi(x, y)$,
which might involve evaluating the model probability $\pmodel(y \mid x)$
as well as collecting human judgments $\hj(x,y)$ from crowdworkers.

Finally, we compute the leave-one-out error of a classifier
that tries to predict whether a given example $(x,y)$
comes from the reference distribution ($z = 1$) or the model ($z = 0$).

The classification problems for HUSE are two-dimensional, which allows us to accurately estimate error rates using a $k$-nearest neighbors classifier. 
We opt to use nearest neighbors classifiers as they are simple, require no training, and can asymptotically capture arbitrary continuous decision boundaries.
Specifically, we set $k=16$
and define neighbors using $L_2$ distances over the feature vectors $\phi(x,y)$ scaled componentwise to have unit variance.
The overall procedure for computing the estimate $\hat{L}(\phi)$ is formally defined in Algorithm~\ref{alg:estimate}.

\begin{algorithm}
  \caption{Estimating error rates under $\phi$}
\label{alg:estimate}
\begin{algorithmic}[1]
  \REQUIRE
  Feature map $\phi$, number of neighbors $k$ \\
  Contexts $x_1, \dots, x_n$ \\
  Reference outputs $y_1, \dots, y_n$ \\
  Model outputs $y_1', \dots, y_n'$ \\
  \STATE Construct dataset:
  \begin{align*}
    \mathcal D &= \bigcup_{i = 1}^n \{ (\phi(x_i, y_i), 1), (\phi(x_i, y_i'), 0) \}
  \end{align*}
  \STATE $\hat{L}(\phi) \eqdef \text{leave-one-out error of $k$-NN on $\mathcal{D}$}$
\end{algorithmic}
\end{algorithm}

\subsection{Quality-diversity decomposition}
\label{sec:moddef}

We now define the (empirical) \emph{HUSE score} using the feature map $\phihuse$:
\begin{align}
\tvhp \defeq \hat{L}(\phihuse).
\end{align}
We define the quality component of HUSE (HUSE-Q) similarly using human judgments alone:
\begin{align}
\tvh \defeq \hat{L}(\phihj).
\end{align}

Since humans can detect quality defects in a model, any increase
in error from removing $\pmodel$ must come from a model's lack of diversity.
Therefore, we define the diversity component (\tvdiv) as follows:
\begin{align}
\tvdiv \defeq 1 + \tvhp - \tvh,
\end{align}
which implies the decomposition $(1 - \tvdiv) + (1 - \tvh) = 1 - \tvhp$.
As long as the discriminators are non-degenerate (obtaining better performance than chance and HUSE $>$ HUSE-Q),
all scores are contained in $[0,1]$.
Here, $\tvdiv = 1$ implies that the model suffers no diversity defects,
while $\tvdiv = 0$ indicates that the examples could be discriminated perfectly due to a lack of
diversity.

\begin{table*}[h!]
\centering
\resizebox{0.95\textwidth}{!}{
\begin{tabular}{c|cc c| cc c| cc c| cc c}
\toprule
\multirow{2}{*}{Score} && \multicolumn{2}{c}{Summarization} && \multicolumn{2}{c}{Story generation} && \multicolumn{2}{c}{Chit-chat dialogue} &&LM \\
&& $t=1.0$  & $t=0.7$  && $t=1.0$ & Retrieval && $t=1.0$ & $t=0.7$ && $t=1.0$\\
\midrule
$\tvhp$ && \textbf{0.53}  & 0.26 && \textbf{0.06} & 0.00 && \textbf{0.56} & 0.49 &&\textbf{0.86 }\\
$\tvh$  && 0.58 & \textbf{0.92} && 0.15 & \textbf{0.47} && 0.56 & \textbf{0.92} && \textbf{0.88}\\
$\tvdiv$ && \textbf{0.95}& 0.34&& \textbf{0.91} & 0.53 && \textbf{1.00} & 0.57 && \textbf{1.02}\\
\bottomrule
\end{tabular}
}
\caption{Performance achieved by the best models on the four tasks, as measured by overall goodness-of-fit ($\tvhp$), sample quality ($\tvh$) and diversity ($\tvdiv$). The scale for HUSE and HUSE-Q ranges from 0.0 (completely distinguishable from reference) to 1.0 (indistinguishable from reference) where the implied classification error is $\tvhp/2$. HUSE-D may exceed 1.0 with small sample sizes when HUSE-Q $>$ HUSE.
  }
  \label{tab:genperf}
\end{table*}

\section{Experiments}
\label{sec:experiments}

\subsection{Experimental setup}

We use $\tvhp$ to evaluate three different types of single-sentence natural language generation tasks:
(i) unconditional and high entropy (language modeling);
(ii) conditional and high entropy (story generation, chit-chat dialogue);
and (iii) conditional and low entropy (summarization). We show that HUSE provides  a direct and interpretable measure of diversity on high-entropy tasks, while also serving as a useful model diagnostic on low-entropy ones.

The four tasks along with the datasets and models are as follows:
\begin{itemize}
\item \textbf{Summarization}: Giganews story to headline dataset and the pre-trained model from \citet{gehrmann2018bottom}. The dataset consists of 3.8 million news story-headline pairs. Examples from this dataset are shown in Table~\ref{tab:examples}.
\item \textbf{Story generation}: Last sentence generation for ROC stories \cite{mostafazadeh2016corpus} consisting of 96,198 examples of partially written four-sentence stories as input, and a single sentence which completes the story as the target. We use a standard OpenNMT model with global attention \cite{klein2017opennmt}.  
\item \textbf{Language modeling}: One billion word benchmark pre-trained language model from \citet{jozefowicz2016exploring}. The task consists of generating a single sentence from the one billion word newswire text distribution.
\item \textbf{Chit-chat dialogue}: Two-turn chit-chat dialogue dataset consisting of 37.3 million comment-response pairs from Reddit (Appendix \ref{sec:reddit}). Comments are generally short (5--15 tokens) and cover a single topic (e.g. given ``wow how did i not notice that'', the response is ``you were focusing on other things its understandable''). We train a convolutional model using fairseq \cite{gehring2017convolutional}.
\end{itemize}

For all the tasks, we train neural models and evaluate their diversity-quality tradeoffs as we change the decoding scheme for generation. Our primary evaluation concerns diversity trade-offs involving \textbf{temperature annealing} which is a generation technique applicable to any probabilistic model that generates words sequentially. In temperature annealed models, we sample a word $w$ proportional to $p^{1/t}(w)$ where $p$ is the model probability of $w$ given previous words and $t$ is the temperature parameter. We excluded beam search since it qualitatively behaves similarly to temperature annealing with low temperatures and $\tvhp \approx 0$ due to beam search being extremely under diverse.

As a non-neural baseline, we also consider retrieval based models based on Apache \texttt{solr} on a few tasks. For this approach, we retrieve the single most relevant response from the training set using the BM25 similarity metric on inputs. Such models are known to perform well in tasks with complex outputs such as program generation \cite{hayati2018retrieval,hashimoto2018edit} and style transfer \cite{li2018style}.

For cost reasons, we did not measure certain combinations of task and generation mechanisms. We did not measure retrieval for chit-chat dialogue, as we observed its outputs were lower quality than a low-temperature neural model. We also did not anneal language models, as the generation quality from the language model was already high, and our goal was to show that they achieved high HUSE. Our set of measurements, while not comprehensive, generally covers the available quality-diversity tradeoffs for conditional tasks.

\begin{figure}
\centering
\includegraphics[scale=0.52]{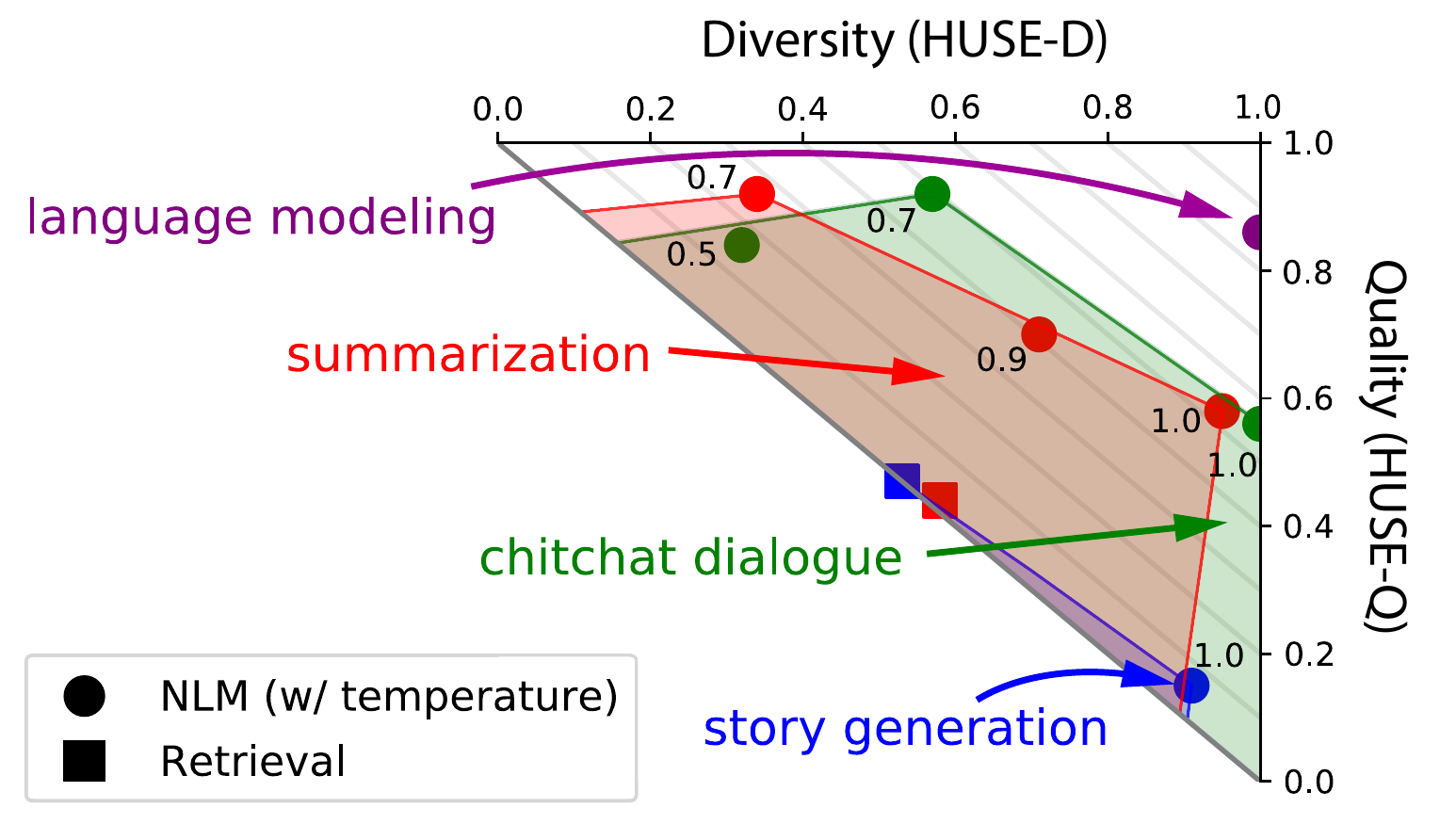}
\caption{Tradeoffs between $\tvdiv$ and $\tvh$. Points are models and color indicates task. Neural models (circle) generate using temperature annealing (point labels indicate temperature). Models closer to the top right are superior, and gray diagonal lines indicate equivalent $\tvhp$.
  A shaded region for a task indicates models which are strictly dominated (worse $\tvhp$ with the same $\tvdiv$-$\tvh$ proportion).
  Annealing can trade-off between diversity and quality but cannot easily increase the underlying model performance ($\tvhp$).
  }
\label{fig:div-qual}
\end{figure}

Finally, we collect human judgments $\hj(x, y)$ as per
Section \ref{sec:estopt} where we query 20 Amazon Mechanical Turk crowdworkers for typicality ratings on 100 reference and 100 model sentences. Since our models generate UNK (unknown and out-of-vocabulary) tokens, we instructed crowdworkers to treat UNK tokens as rare, but appropriate words for the context.

\subsection{Overall results}

The $\tvhp$ scores across the four tasks vary widely. Table \ref{tab:genperf} shows that single-sentence language models are nearly indistinguishable, with $\tvhp=0.86$ and implied discriminator error of $43\%$.

In contrast, both summarization and dialogue are highly distinguishable ($\tvhp \approx 0.5$) with relatively low quality when sampled from $t=1.0$. Human evaluation alone (HUSE-Q) would suggest that using temperature annealing $(t=0.7)$ to emphasize high-probability outputs substantially improves the model (HUSE-Q goes from $0.58$ to $0.92$ for summarization and $0.56$ to $0.92$ for dialogue). However, we find that this increase in sample quality comes at the cost of diversity (HUSE-D goes from $0.95$ to $0.34$ for summarization and $1.0$ to $0.57$ for dialogue).
Examining the achievable $\tvhp$ and diversity tradeoffs in Figure \ref{fig:div-qual} shows that mechanisms such as annealing which improve sample quality actually degrade $\tvhp$ due to severe losses in diversity.

We find that all generation schemes and models are inadequate for story generation
on ROC stories. The original model ($t=1.0$) is very easily distinguishable
by a human ($\tvh=0.15$), corresponding to a discriminator error of
$7\%$. The retrieval models can improve this to $\tvh=0.47$, but this comes at
the expense of diversity. 

Finally, we observe that directly sampling from the model $(t=1.0)$ is always diverse. This suggests that human evaluation is an appropriate evaluation for generation systems that are directly sampled (rather than beam-searched). %

\begin{figure*}[h!]
\includegraphics[scale=0.45]{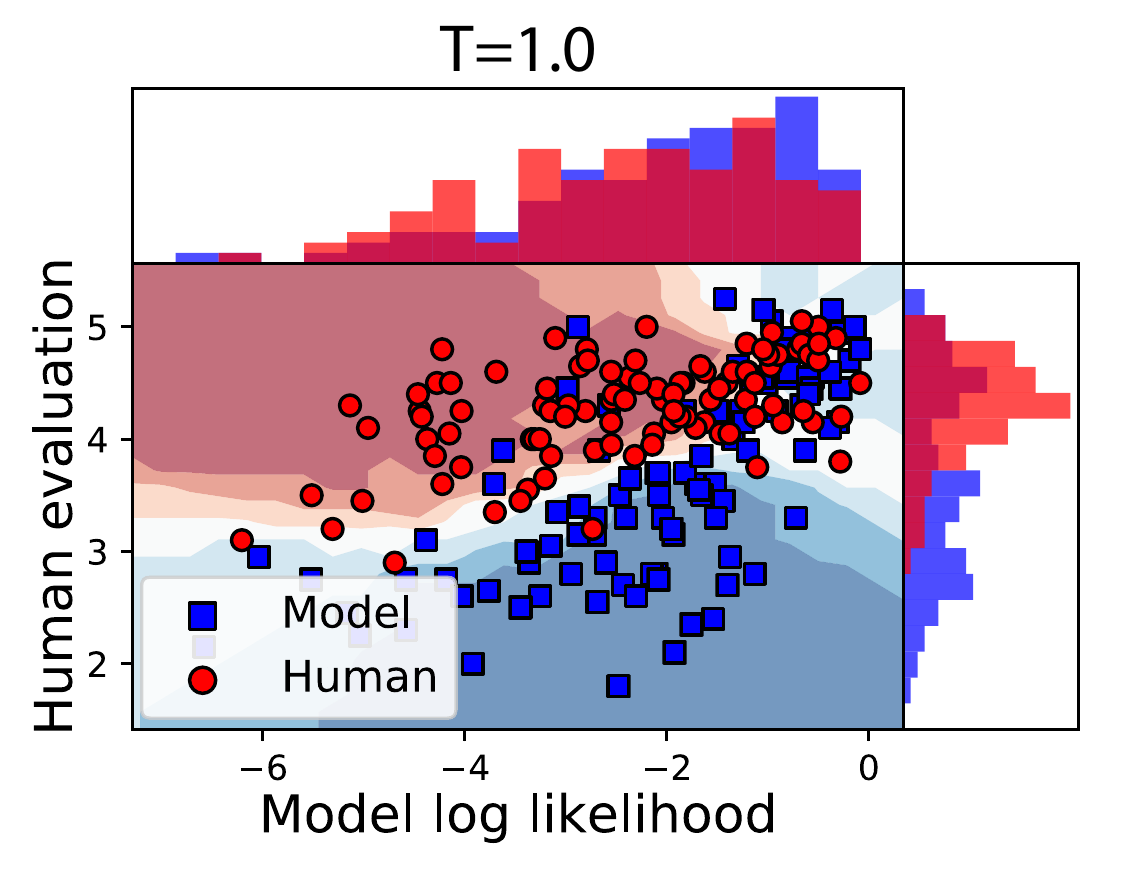}
\includegraphics[scale=0.45]{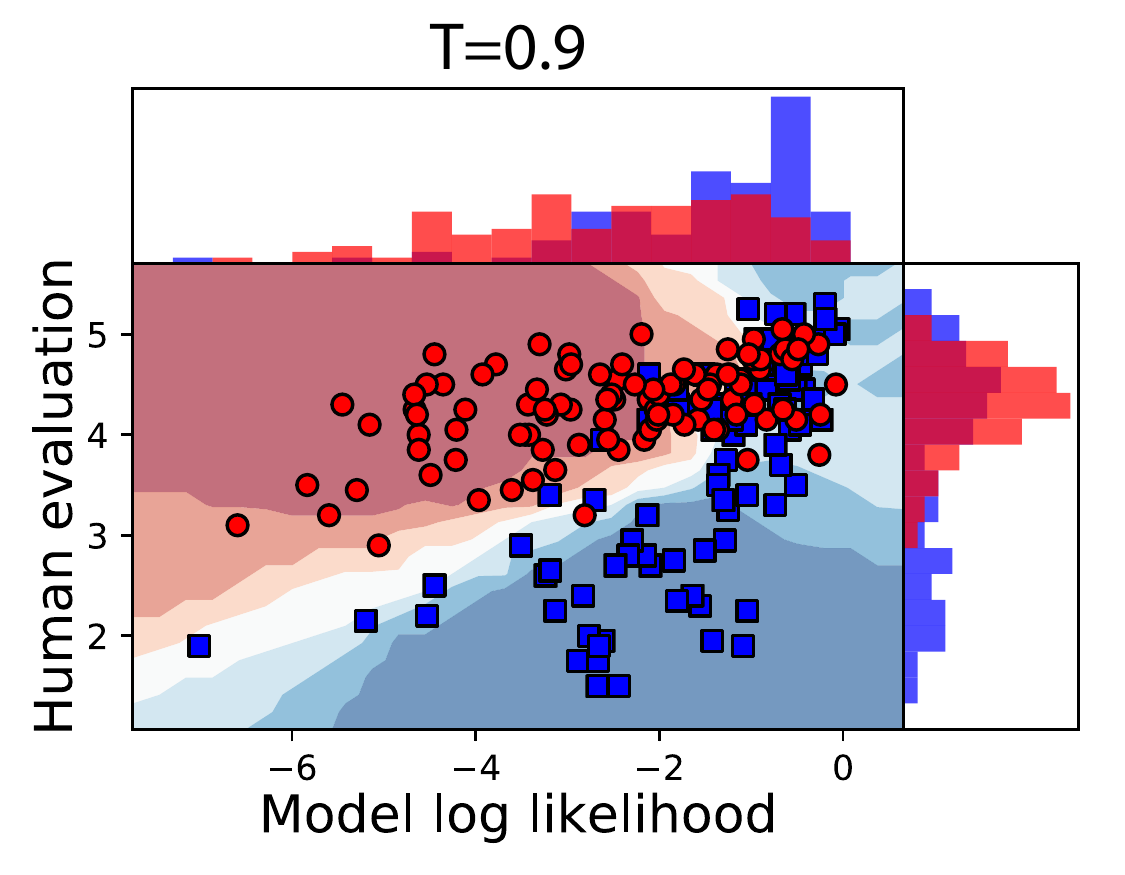}
\includegraphics[scale=0.45]{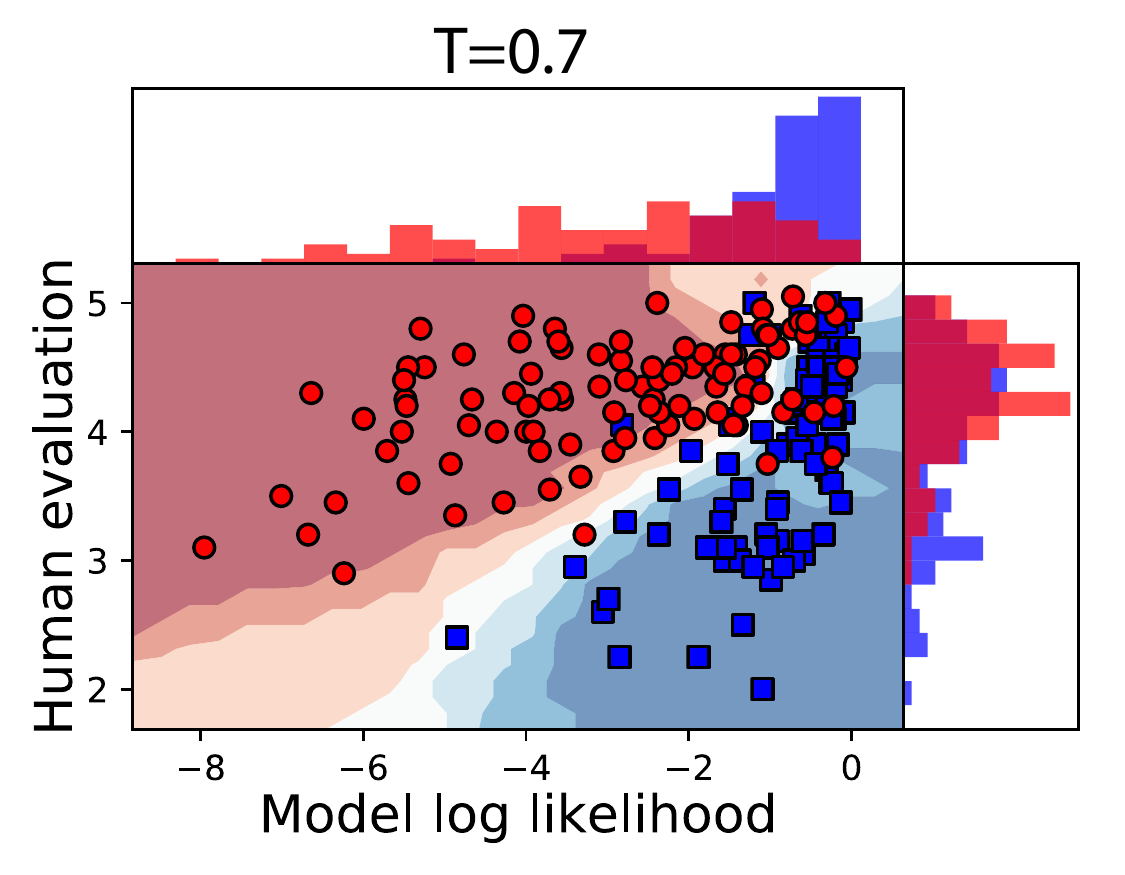}
\caption{The two-dimensional classification problem in Algorithm \ref{alg:estimate} on the summarization task with different softmax temperatures (three panels).
  Each point represents a reference sentence $\phihuse(x_i,y_i)$ or model-generated sentence $\phihuse(x_i,y_i')$.
  The color denotes the source of the sentence ($z$), shading is the classification confidence of the nearest neighbor classifier.
  }
\label{fig:sumvis}
\end{figure*}

\subsection{Model error analysis with $\tvhp$}

Since $\tvhp$ is estimated from a two-dimensional classification problem, we
can directly visualize the classification problem to understand defects in both model quality and diversity. 
  
Figure \ref{fig:sumvis} shows
both reference points $\phihuse(x_i, y_i)$ (blue squares) and model points $\phihuse(x_i, y_i')$ (red circles)
for the summarization task.
The shaded areas indicate the decision boundary of the $16$-nearest
neighbor classifier.

At temperature $t=1.0$, we find that the classification boundary is mostly
horizontal, implying that human judgment alone can distinguish model outputs
from references.
There is a cluster of sentences with high $\hj$ and high $\pmodel$
which are essentially indistinguishable. Examining the samples in this
top-right region reveals that these are news stories with short headlines such
as ``Nadal pulls out of Sydney International'' which can be reliably generated
even at $t=1.0$. However, the model frequently generates low quality samples that can easily be distinguished such as ``two new vaccines in the poor countries were effective against go-it-alone study says'' (Table \ref{tab:examples}).

At lower temperatures of $t=0.9$ and $t=0.7$, the boundary shifts towards
becoming diagonal. Although the distribution is no longer directly separable on
human judgment, the two distributions are clearly separable with the inclusion of
$\pmodel$.

Using Figure~\ref{fig:sumvis}, we can identify individual examples which were correctly
and incorrectly classified based on $\pmodel$ and $\hj$. Table
\ref{tab:examples} shows examples of both quality failures and diversity
failures identified by $\tvhp$. For example, the ``diversity failure'' table
shows that the summarization model ($t=0.7$) has an extremely low probability of generating some reference sentences (``NFL’s bills shake up front office'') and is thus under-diverse. Closer examination of the model shows that the probability of generating ``front office'' is low, since it is an unusual way to refer to the president and general manager. Improving these models on the diversity failures
will require that the model understand more subtle paraphrases.
We can also identify model successes,
where the model outputs are indistinguishable from the reference in terms of quality (``Agassi bows out of Australian Open after injury''), and the model assigns high probability to the reference (``Agassi withdraws from Australian Open'').

\begin{table*}[!t]
  \newcommand{\unk}{\text{$<$UNK$>$}\:}
\centering \small
\resizebox{.95\textwidth}{!}{
\begin{tabularx}{\linewidth}{l X c c  }
\multicolumn{2}{c}{\textbf{\emph{Quality failure}}} & $\log \pmodel$ & $\hj$ \\
\toprule
\textbf{Context:} & Two new vaccines have been shown effective against rotavirus, which is responsible for a half-million infant deaths in poor countries each year, research studies published Wednesday said. & &\\
\midrule
\textbf{Model} & Two new vaccines in the poor countries were effective against go-it-alone study says &-2.3 & 2.6  \\
\midrule
\textbf{Reference} & New vaccines for key \unk virus shown effective & -4.0 & 4.3  \\

  ~\\
\multicolumn{2}{c}{\textbf{\emph{Diversity failure}}} & & \\
\toprule
\textbf{Context:} & The Buffalo Bills sacked Tom Donahoe as president and general manager on Wednesday, fulfilling expectations of a shake-up after another failure to make the National Football League playoffs. & &  \\
\midrule
\textbf{Model} & Bills sack \unk as president GM and general manager & -0.9 & 4.3  \\
\midrule
\textbf{Reference} & NFL's Bills shake up front office. & -5.1 & 4.3 \\

~\\
\multicolumn{2}{c}{\textbf{\emph{Model is indistinguishable}}} & & \\
\toprule
\textbf{Context:} & US veteran and eight-time Grand Slam winner Andre Agassi has withdrawn from this month's Australian Open due to a nagging ankle injury, his management team announced Thursday. & & \\
\midrule
\textbf{Model} & Agassi bows out of Australian Open after injury. & -1.4 & 5.3 \\
\midrule
\textbf{Reference} & Agassi withdraws from Australian Open. & -0.3 & 4.9 \\
\end{tabularx}
}
\vspace{8pt}
\caption{
  Example reference and model outputs (capitalization added for readability) corresponding to Figure \ref{fig:sumvis} (summarization task) that were shown to crowdworkers (left column). Crowdworkers were shown samples from the model (including the \unk token) and returned human judgments (right column).
  Using human judgments and the model probability, we can identify several types of failures. Quality failures are examples that are classified by human judgment. Diversity failures are examples that are classified by model probabilities. Finally some examples are not easily classified, as they have similar human judgment and model probability scores.
}
\label{tab:examples}

\end{table*}

\subsection{HUSE stability}
\label{sec:exptdesign}

\begin{figure}

\includegraphics[scale=0.62]{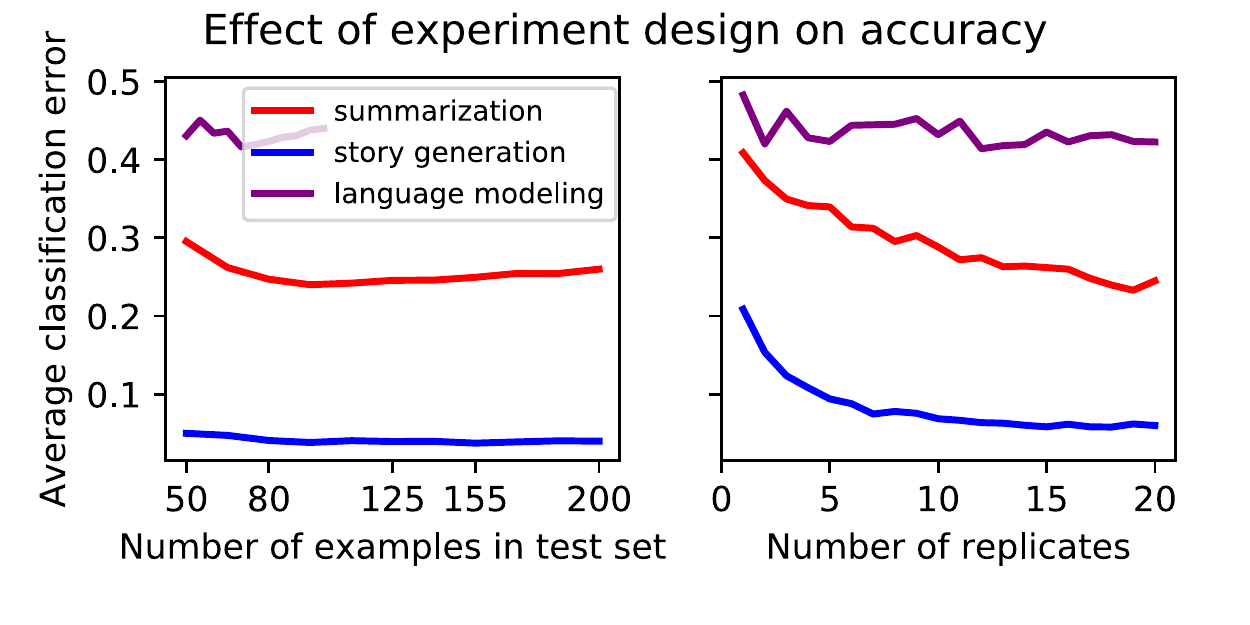}
\caption{Estimates of $\tvhp$ are robust to small test set size, but generally require $\approx 20$ crowdworker measurements for each example.
  }
  \label{fig:exptdesign}
\end{figure}

Since HUSE depends on human crowdworker annotations, one might ask if it is possible to reduce either the number of annotated examples, or number of distinct crowdworkers for each example. We show that for low-quality models, substantially fewer annotations are needed.

Figure \ref{fig:exptdesign} shows the result of subsampling our original data of 200 sentences and 20 crowdworkers and estimating $\tvhp$. 
First, we find that using 50 test set examples (Figure \ref{fig:exptdesign}, left) is often sufficient to give accurate estimates of HUSE. 
Next, we find that the necessary number of crowdworkers per example depends heavily on the task. Easily
distinguishable tasks (story generation), require only 10 crowdworkers, while less
distinguishable tasks (summarization) require more than 20 crowdworkers to obtain
accurate estimates.

\section{Related work}
\label{sec:related}

\paragraph{The current state of NLG evaluation.}

Existing approaches to NLG evaluation use a hodgepodge mix of quality and diversity
measures. Out of the 26 NLG papers at ACL 2018, six perform only
human evaluation, fourteen measure human evaluation and a diversity metric such
as perplexity or n-gram diversity, and six do not evaluate using human
judgments.

While perplexity and $n$-gram counts can \emph{in principle} evaluate diversity, their
practical implementations suffer from serious drawbacks.
When human evaluation and perplexity are both evaluated, they are almost always done on separate
models---human evaluations are done on beam-searched output, while perplexity
is computed on the softmax outputs. This makes it appear as if the models can
simultaneously generate high quality outputs while also being diverse, when in
fact they can only be one at a time based on whether they sample or
run beam search.

On the other hand, $n$-gram diversity was proposed by \citet{li2016diversity} to
identify models with the generic utterance problem where models repeat phrases such as `I
don't know'. Unfortunately, $n$-gram diversity is computed \emph{across}
contexts by counting the number of unique $n$-grams generated, and so does not
measure a model's ability to generate multiple valid utterances at any single
context. In particular, a model which only outputs a single memorized utterance per
context (e.g., via memorization or retrieval) can still have high $n$-gram
diversity as long as the memorized sentences differ across contexts.

Finally, \emph{all} existing diversity measures are computed separately from
human evaluation. This results in two incomparable evaluation metrics, which prevent us from reasoning about tradeoffs between diversity and quality. In contrast, HUSE allows us to make precise statements about the tradeoffs between model quality and diversity because it is a single metric which decomposes into diversity and quality terms.

\paragraph{Related evaluations of diversity.}

The importance of diverse responses has previously been acknowledged for
summarization \cite{nenkova2007pyramid} and information retrieval
\cite{clarke2008diversity}. Our work differs in considering a \emph{single}
evaluation measure that captures quality and diversity applicable to \emph{any}
generation task.

Automated metrics based on $n$-gram overlap such as BLEU, METEOR, ROUGE
\cite{papineni02bleu,lavie2009meteor,lin2004rouge} work well for machine
translation but do not generalize well to domains with a diverse spectrum of
correct responses. While variants \cite{sun2012joint, galley2015delta,
shima2011diversity} have adapted such metrics to high entropy generative
environments, they are still significantly inferior to the human judgments
they attempt to mimic.

\citet{languagegans2018} recently examined the diversity and quality tradeoffs for different language model architectures on synthetic datasets.
However, as their approach relies on measuring log-likelihoods under both the model and reference distributions,
it \emph{cannot} be applied to real data where $\phum$ is unavailable.
Our main conceptual contribution overcomes this by showing that $\hj$ is an acceptable proxy for $\phum$. %

\citet{sajjadi2018precision} also examines diversity and quality (which they call precision and recall) in the context of generative image models. However, they rely on assuming that $\phum$ and $\pmodel$ can be estimated accurately using the Fréchet Inception Distance (FID) \cite{huesel2017gans}. %
HUSE avoids such assumptions and instead directly leverages human judgments, resulting in a simple and reliable metric more suitable for use as a gold-standard.

\paragraph{Estimating optimal classification error.}

Evaluating a model by estimating its optimal classification error has been considered by several earlier works \cite{olsson2018skill, kannan2016adversarial, li2017adversarial, bruni2017adversarial, bowman2016continuous}. However, these methods have focused on classifying sentences directly, which is quite challenging to do reliably. Existing adversarial evaluation methods do not yet reliably outperform human classification \cite{kannan2016adversarial, bruni2017adversarial}. We propose the use of both human evaluation and model probabilities as part of the adversarial evaluation framework, and demonstrate that the resulting classifier reliably outperforms humans and captures both the sample quality and diversity of a model.

\paragraph*{Distributional divergence estimation.}
Our proposed evaluation metric is closely
related to the total variation distance which has been studied
extensively in the distribution
testing literature. It is known that total variation distance estimates have pessimistic 
minimax estimation rates in high dimensions \cite{balakrishnan2017hypothesis}.
Our work overcomes this by utilizing $\pmodel$ and an estimate of $\phum $.
Other approaches to distributional testing include the
maximum mean discrepancy (MMD) and Wasserstein distances, but these approaches require knowledge of a ground truth metric or kernel space \cite{tolstikhin2016minimax,singh2018nonparametric}. Although such
divergences are easier to estimate than the total variation distance from samples, the
implied convergence rates are still too slow to be practically useful.

\section{Discussion}

In this paper, we demonstrate that the current gold standard of human evaluation does not penalize under-diverse models. To remedy this, we propose HUSE, a general purpose evaluation strategy which can be applied to any model for which we can calculate a model's sampling probabilities. HUSE is an upper bound on the optimal classification error of distinguishing reference and model-generated text, and  \emph{never} does worse than human classification. HUSE leverages both model probabilities and human judgments, ensuring that models which do well on the metric are both high-quality and diverse. 

Our work can be viewed as a ``superhuman version'' of the classic Turing Test \cite{turing1950computing}.
Instead of relying on just a human classifier,
we approximate the \emph{optimal} classifier, which can utilize information about the model in addition to the reference. We also modify the classification problem and seek to identify whether a sample comes from a (potentially superhuman) reference distribution, rather than the human distribution. These two changes lead to tractable, rigorous estimators which can quantify tradeoffs between model quality and diversity on a wide range of generation tasks.

\textbf{Acknowledgements.} We would like to thank Arun Chaganty, Robin Jia, and Peng Qi for extensive comments and feedback on the paper. This work was funded by DARPA CwC program under ARO prime contract no. W911NF-15-1-0462. 

\textbf{Reproducibility.} All code, data, and experiments are available on the CodaLab platform at \url{https://worksheets.codalab.org/worksheets/0x88644b5ee189402eb19d39d721d1005c}.

\bibliography{main.bbl}
\bibliographystyle{acl_natbib}

\clearpage
\appendix
\section{Appendix}
\label{sec:appendix}

\subsection{Relationship between total variation distance and optimal discriminator error}
\label{sec:tverr}

This is a standard result, replicated here for completeness:
\begin{proposition}
The total variation distance is related to the optimal discriminator error as follows:
$\|\pmodel - \phum\|_\text{TV} = 1 - \optclass$.
\end{proposition}
\begin{proof}
Fix any $x$.
Define $a_y \defeq \phum(y \mid x)$ and $b_y \defeq \pmodel(y \mid x)$.
Let $S \defeq \{ y : a_y < b_y \}$ be the $y$ where the $\pmodel$ assigns higher probability than $\phum$,
and define $A \defeq \sum_{y \in S} a_y$ and $B \defeq \sum_{y \in S} b_y$ be the aggregated probabilities.
On $S$, the optimal discriminator should return $z=0$ (model).
This is an error when $z = 1$, which occurs with probability $\frac12 A$.
  Analogously, on the complement of $S$, the error probability (when $z = 0$) is $\frac12 (1 - B)$.
  The total contribution to $\optclass$ is thus $A + (1 - B)$.
The rest follows from algebra:
  \begin{align}
  & \|\pmodel - \phum\|_\text{TV} = \frac{1}{2} \| \pmodel - \phum \|_1 \\ 
  =\  & \frac{1}{2} [(B - A) + (1-A) - (1-B)] \\
  =\  & B - A = (1 - \optclass).
  \end{align}
\end{proof}

\subsection{Approximation error from $\phi$ features}

\begin{thm}\label{thm:approxerr}
  Let $L^*$ and $L(\phi)$ be the optimal classification error and optimal error under feature map $\phi$ respectively. Then,
  \[L^* \leq L(\phi) \leq L^* + 2(1-2^{-I}) \]
  where $I\defeq I(Z_{opt};\phi_{opt}(X,Y)\mid \phi(X,Y))$ is the conditional mutual information in bits and $Z_{opt}$ is the prediction of the optimal classifier.
\end{thm}
\begin{proof}
  The lower bound falls out of the definition of $L^*$.
  To prove the upper bound, a variant of the entropy lower bound by Feder and Merhav \cite{feder1994relations} shows that the error rate for predicting $Z_{opt}$, via the optimal $f(\phi(X,Y))$ follows
  \begin{multline}\label{eq:errbound}
    P(f(\phi(X,Y)) \neq Z_{opt} ) \\
    \leq 1-2^{I(Z_{opt} ; \phi(X,Y)) - H(Z_{opt})}.
  \end{multline}
  Now expand the mutual information using the chain rule
  \begin{align*}
    &I(Z_{opt} ; \phi(X,Y))  = I(Z_{opt} ; \phi_{opt}(X,Y) , \phi(X,Y))\\
    &\qquad  -I(Z_{opt} ; \phi_{opt}(X,Y) \mid \phi(X,Y))\\
    &\quad= -I(Z_{opt} ; \phi_{opt}(X,Y) \mid \phi(X,Y)) + H(Z_{opt}).\\
  \end{align*}
  The last line follows from the fact that $Z_{opt}$ is a deterministic function of $\phiopt$ (Proposition \ref{prop:phiopt}). Substituting this into the inequality gives the bound,
  \[P(f(\phi(X,Y)) \neq Z_{opt} ) \leq 1-2^{-I}\]
  with $I= I(Z_{opt};\phi_{opt}(X,Y)\mid \phi(X,Y))$.

  Finally, note that $Z_{opt}$ incurs $L^*/2$ error, and we disagree with $Z_{opt}$ at most a $P(f(\phi(X,Y)) \neq Z_{opt} )$ fraction of time. Assuming that we get every one of these disagreements wrong gives an upper bound of $L^*/2 + P(f(\phi(X,Y)) \neq Z_{opt} )$ on $L(\phi)/2$.
\end{proof}

A straightforward corollary is that whenever $\phi$ is an invertible function of $\phiopt$, the conditional mutual information is zero, and therefore the above inequalities become an equality.
\begin{corollary}\label{cor:invertexact}
  Whenever $\phi$ is an invertible function of $\phiopt$, $L(\phi) = L^*$.
  \end{corollary}

\subsection{Amazon Mechanical Turk for human judgments}
\label{sec:amt}

\begin{figure*}[ht!]
\centering
\includegraphics[scale=0.6]{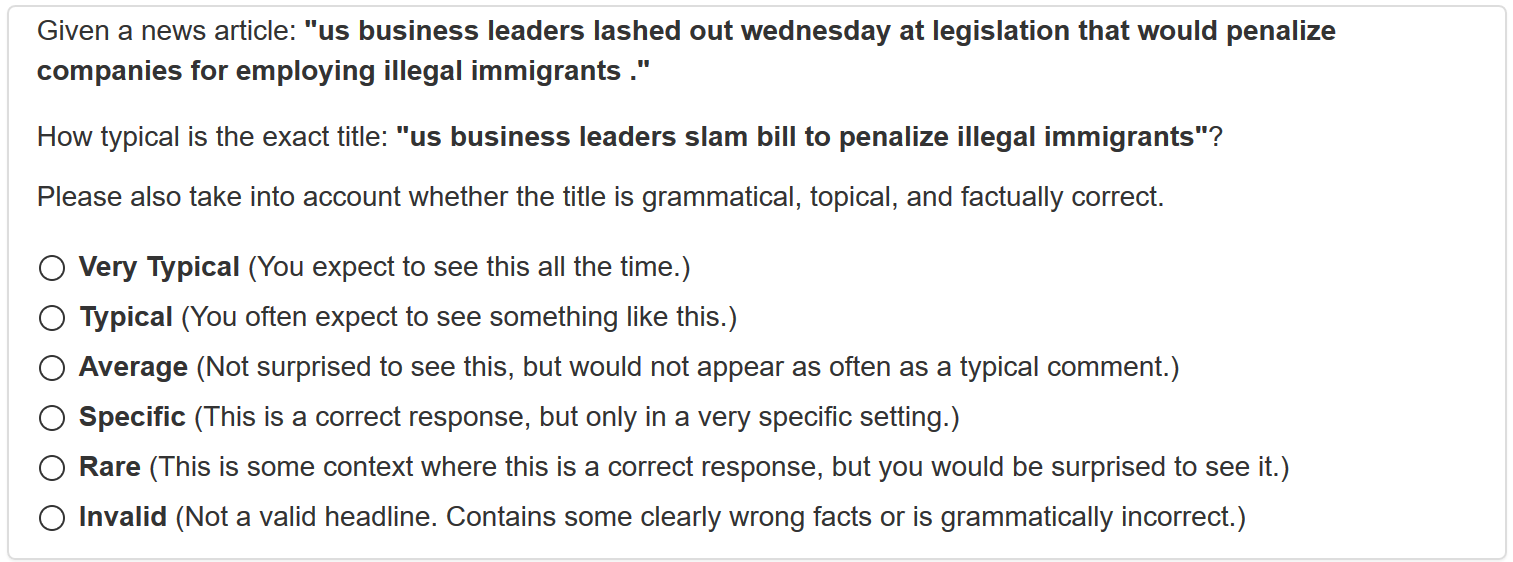}
\caption{Amazon Mechanical Turk survey design for eliciting human judgment scores $\hj$ in the summarization task.}
\label{fig:turkex}
\end{figure*}

In order to show that $\tvhp$ can be reliably estimated even with simple crowdsourcing techniques, we used a single uniform task design where we asked Amazon Mechanical Turk workers to rate the typicality of a sentence from 0--5. We defined 0 as invalid (grammatically or factually incorrect) and 5 as `very typical'. $\hj(x, y)$ is defined as the average score that crowdworkers assign to a response $y$ given the context $x$. We did not perform substantial filtering or qualification checks beyond HIT acceptance rate (HIT Approval rate greater than 95 percent and number of HITs approved greater than 50 and location is USA).
We constructed each HIT to be 25 examples, and paid one dollar per HIT. 

We observe that measuring many replicates is sufficient to get low-variance estimates of $\hj$.
For classification tasks where the model is straightforward to identify from references (such as story generation) we require five to ten replicates,
while for hard tasks such as summarization at least twenty replicates are needed (Section \ref{sec:exptdesign}).
Manual inspection suggests that up to 20\% of the collected data are low-quality
but that this noise is uncorrelated with the sentence being rated and
outweighed by a larger majority of honest and reasonably accurate data. Even if the data quality is low, HUSE is still a valid upper bound (i.e. models with low HUSE are guaranteed to be distinguishable from humans). Thus the models which we identify as having low-HUSE are reliably distinguishable regardless of the crowdworker quality.

\subsection{Reddit Dataset}
\label{sec:reddit}

We use a subset of Reddit comments from 2006-2018 scraped from https://pushshift.io/. We construct a dictionary containing the 10,000 most popular words and preprocess the dataset by removing deleted posts, out-of-vocabulary tokens, profanity, comments with less than 10 upvotes, and comments with over 400 tokens.

\end{document}